\documentclass[10pt,twocolumn,letterpaper]{article}

\usepackage{cvpr}
\usepackage{times}
\usepackage{epsfig}
\usepackage{graphicx}
\usepackage{amsmath}
\usepackage{amssymb}

\usepackage{dsfont}
\usepackage{siunitx}
\usepackage{subfigure}
\usepackage{multirow}
\usepackage{caption}
\usepackage{booktabs}       
\usepackage{standalone}
\usepackage{xspace}
\usepackage{comment}
\usepackage{multirow}

\usepackage[breaklinks=true,bookmarks=false]{hyperref}

\newcommand{\myparagraph}[1]{\vspace{4pt}\noindent{\bf #1}}

\cvprfinalcopy 

\def\cvprPaperID{4838} 

\ifcvprfinal\fi
\begin{document}

\title{Grounding Human-to-Vehicle Advice for Self-driving Vehicles} %
\author{Jinkyu Kim$^{1}$, Teruhisa Misu$^{2}$, Yi-Ting Chen$^{2}$, Ashish Tawari$^{2}$, and John Canny$^{1}$\\
$^{1}$EECS, UC Berkeley, $^{2}$Honda Research Institute USA, Inc.\\
{\tt\small $^{1}$\{jinkyu.kim, canny\}@berkeley.edu, $^{2}$\{tmisu,ychen,atawari\}@honda-ri.com}
}

\maketitle
\thispagestyle{empty}

\begin{abstract}
   Recent success suggests that deep neural control networks are likely to be a key component of self-driving vehicles. These networks are trained on large datasets to imitate human actions, but they lack semantic understanding of image contents. This makes them brittle and potentially unsafe in situations that do not match training data. Here, we propose to address this issue by augmenting training data with natural language advice from a human. Advice includes guidance about what to do and where to attend. We present a first step toward advice giving, where we train an end-to-end vehicle controller that accepts advice. The controller adapts the way it attends to the scene (visual attention) and the control (steering and speed). Attention mechanisms tie controller behavior to salient objects in the advice. We evaluate our model on a novel advisable driving dataset with manually annotated human-to-vehicle advice called Honda Research Institute-Advice Dataset (HAD). We show that taking advice improves the performance of the end-to-end network, while the network cues on a variety of visual features that are provided by advice. The dataset is available at https://usa.honda-ri.com/HAD.
\end{abstract}


\section{Introduction}
Dramatic progress in self-driving vehicle control has been made in the last several years. The recent achievements~\cite{bojarski2016end, xu2016end} suggest that deep neural models can be applied to vehicle controls in an end-to-end manner by effectively learning latent representations from data. Explainability of these deep controllers has increasingly been explored via a visual attention mechanism~\cite{kim2017interpretable}, a deconvolution-style approach~\cite{bojarski2016visualbackprop}, and a natural language model~\cite{kim2018textual}. Such explainable models will be an important element of human-vehicle interaction because they allow people and vehicles to understand and anticipate each other's actions, hence to cooperate effectively.

\begin{figure}[!t]
    \begin{center}
        \includegraphics[width=\linewidth]{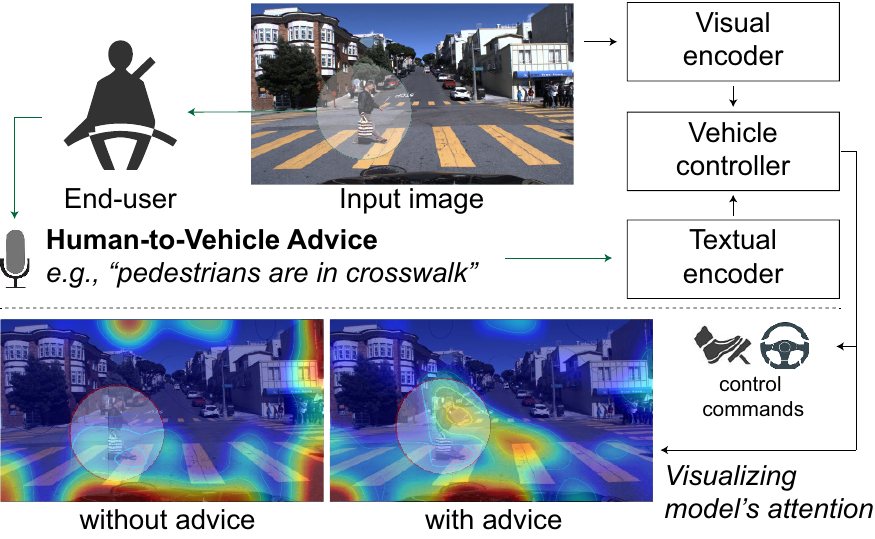}
    \end{center}
    \caption{Our model takes human-to-vehicle advice as an input, i.e., ``pedestrians are in crosswalk'', and grounds it into the vehicle controller, which then predicts a sequence of control commands, i.e., a steering wheel angle and a vehicle's speed. Our driving model also provides a visual explanation in the form of attention - highlighted regions have a direct influence on the function being estimated. Visualizing attention maps helps the end-users acknowledge the acceptance of their advice.}
    \label{fig:overview}\vspace{-1em}
\end{figure}

However, the network's understanding of a scene is limited by the training data: image areas are only attended to if they are salient to the (training) driver's subsequent action. We have found that this leads to semantically-shallow models that under-attend to important cues (like pedestrians) that do not predict vehicle behavior as well as other cues, like the presence of a stop light or intersection. We also believe its important for driving models to be able to adapt the ``style'' of the journey to user input (fast, gentle, scenic route, avoid freeways etc). We
use the term ``advice'' to cover high-level instructions to the vehicle controller about how to drive, including what to attend to. We distinguish advice from explicit commands to the vehicle: which may be problematic if the passenger is not fully attending to the vehicle's environment. 

The goal of this work is to augment imitation learning datasets with long-term advice from humans (e.g., driving instructors) and in the shorter term, from passengers in the vehicle. In full generality, advice might take the form of condition-action rules ``if you see a child's toy on the sidewalk, slow down''. For the present paper, we study the simpler task of accepting short-term textual advice about action or perception.  
 
In this work, we propose a novel driving model that takes natural language inputs (i.e., human-to-vehicle advice) from an end-user. Here, we focus on two forms of advice: (1) goal-oriented advice (top-down signal) -- to influence the vehicle in a navigation task (e.g., ``drive slow in a school zone''), (2) stimulus-driven advice (bottom-up signal) -- conveys some visual stimuli that the user expects their attention to be actively looked by the vehicle controller (e.g., ``there is a pedestrian crossing''). As shown in Figure~\ref{fig:overview}, the controller needs three main capabilities to handle such advice. (i) Perceptual primitives to evaluate the controller's behavior. (ii) The ability to understand the user's utterance and to ground it in the trained perceptual primitives. (iii) Explainability of the controller's internal state to communicate with the vehicle. We propose that such capabilities can be learned during off-line training.

Our contributions are as follows. (1) We propose a novel advisable driving model that takes human-to-vehicle advice and grounds it into the vehicle controller. (2) We internalize the (stimulus-driven) advice -- aligning its attention to make the model refer to the important salient objects even when advice is not available. (3) We generated a large-scale dataset called Honda Research Institute-Advice Dataset (HAD) with over 5,600 video clips (over 32 hours) with human-to-vehicle advice annotations, e.g., ``there is a pedestrian pushing a stroller through the crosswalk''. The dataset will be available and will provide a new test-bed for measuring progress towards developing advisable models for self-driving cars.

\section{Related Work}
\myparagraph{End-to-End Learning for Self-driving Vehicles.}
Recent successes~\cite{bojarski2016end,xu2016end} suggest that a driving policy can be successfully learned by neural networks as a supervised learner over observation (i.e., raw images)-action (i.e., steering) pairs collected from human demonstration. Bojarski~\etal~\cite{bojarski2016end} trained a deep neural network to map a dashcam image to steering controls, while Xu~\etal\cite{xu2016end} explored a stateful model using a dilated deep neural network and recurrent neural network so as to predict a vehicle's discretized future motion given input images. Other variants of deep neural architecture have been explored~\cite{fernando2017going, chi2017learning}.

Explainability of deep neural networks has become a growing field in computer vision and machine learning communities. Kim~\etal\cite{kim2017interpretable} utilized a recurrent attention model followed by a causal filtering that removes spurious attention blobs and visualizes causal attention maps. We start our work with this attention-based driving model. Attention model visualizes controller's internal state by visualizing attention maps, which end-users may use as a ground and an acknowledgment of their advice. Other approaches~\cite{bojarski2016visualbackprop, kim2018textual} can also be applied to provide richer explanations, but we leave it for future work. 

\myparagraph{Advice-taking models.}
Recognition of the value of advice-taking has a long history in AI community~\cite{mccarthy1960programs}, but a few attempts have been made to exploit textual advice. Several approaches have been proposed to translate the natural language advice in formal semantic representations, which is then used to bias actions for simulated soccer task~\cite{kuhlmann2004guiding}, mobile manipulation tasks~\cite{misra2016tell, misra2015environment, tellex2011understanding}, and a navigation task~\cite{artzi2013weakly}. These approaches consider high-level action sequences to be given in the task space of the agent. Instead, we consider the visual imitation learning setting, where the model has its own perceptual primitives that are trained by observing third-person demonstration and types of advice. Recent work suggests that incorporation of natural language human feedback can improve a text-based QA agent~\cite{li2016dialogue, weston2016dialog} and image captioning task~\cite{ling2017teaching}. Despite their potential, there are various challenges (e.g., safety and liability) with collecting human feedback on the actions taken by self-driving cars. Other notable approaches (in the reinforcement learning setting) may include the work by Tung~\etal~\cite{tung2018reward} that learns a visual reward detector conditioned on natural language action descriptions, which is then used to train agents. To our best knowledge, ours is the first attempt to take human-to-vehicle advice in natural language and ground it in a real-time deep vehicle controller. 

\section{Advisable Driving Model}
\begin{figure*}[!t]
    \begin{center}
        \includegraphics[width=\linewidth]{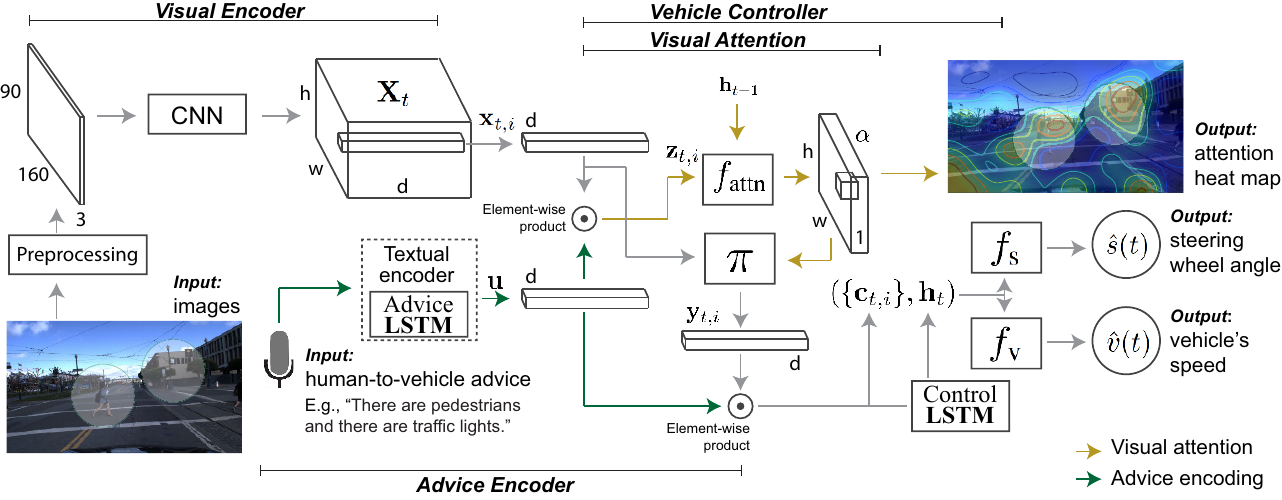}
    \end{center}
    \caption{Our model consists of three main parts: (1) a visual encoder (CNN here), (2) an advice encoder, which encodes end-user's utterance (advice) and ground it into the vehicle controller (see green arrows), and (3) an interpretable vehicle controller, which predicts two vehicle control commands (i.e., a speed and a steering angle command) from an input raw image stream in an end-to-end manner. Our model also utilizes a (spatial) visual attention mechanism to visualize where and what the model sees (see yellow arrows). }
    \label{fig:model}\vspace{-1em}
\end{figure*}

As we summarized in Figure~\ref{fig:model}, our model involves three main parts: (1) a {\em{Visual encoder}}, which extract high-level visual descriptions by utilizing the convolutional neural network (CNN). (2) An {\em{Advice encoder}}, which is a natural language model that encodes end-user's utterance into a latent vector and ground it into the vehicle controller. (3) An {\em{Interpretable vehicle controller}}, which is trained to predict two control commands (widely used for self-driving vehicle control) in an end-to-end manner, i.e., a vehicle's speed and a steering wheel angle. Our controller uses a visual (spatial) attention mechanism~\cite{kim2017interpretable}, which visualizes controller's internal state by highlighting image regions where the model fixates on for the network's output.

\subsection{Preprocessing}
Following \cite{kim2017interpretable}, we use raw images that are down-sampled to 10Hz and are resized to have input dimensionality as 90$\times$160$\times$3. For better generalization, each image is then normalized by subtracting its mean from the raw pixels and dividing by its standard deviation. Following Liu~\etal~\cite{liu2016ssd}, we marginally change its saturation, hue, and brightness for achieving robustness during a training phase.

\subsection{Convolutional Feature Encoder}
We utilize a convolutional neural network (CNN) to obtain a set of visually-descriptive latent vectors at time $t$, where each vector contains a high-level visual description in certain input region. In this paper, we refer these latent vectors to as a convolutional feature cube ${\bf{X}}_t$. By feeding an image through the model at each time $t$, we collect a ${\bf{X}}_t$ of size $w$$\times$$h$$\times$$d$. Note that ${\bf{X}}_{t}$ has $l$ (=$w$$\times$$h$) (spatially) different vectors, each of which is a $d$-dimensional feature slice corresponding to a certain input region. Choosing a subset of these vectors will allow us to focus selectively on different parts of images (i.e., attention). Formally, ${\bf{X}}_{t}$ = $\{$${\bf{x}}_{t,1}$, ${\bf{x}}_{t,2}$, $\ldots$, ${\bf{x}}_{t,l}\}$, where ${\bf{x}}_{t,i}\in{\cal{R}}^{d}$ for $i\in\{1,2,\ldots,l\}$.

\subsection{Advice Encoder}
Our advice encoder takes a variable-length advice and yields a latent vector, which then feeds to the controller LSTM (called Control LSTM). Our advice-taking driving model needs to understand the end-users utterance and to ground it into the vehicle controller. We assume that advice will often be given offline, or at the beginning of a ride, e.g., ``look out for pedestrians'' or ``drive gently (occupant gets carsick)''. Thus, advice encoding will be prepared ahead of the controller generates control commands. 

We train our advice encoder to deal with both types of advice (i.e., the goal-oriented and the stimulus-driven advice) without any input-level separation. We use a LSTM (called Advice LSTM, which is different from Control LSTM) to encode an input sentence (i.e., human-to-vehicle advice) and to yield a fixed-size latent vector, which is common practice in sequence-to-sequence models. Inspired by the knowledge from the Visual Question Answering (VQA) task, we follow the work by Park~\etal~\cite{park2018multimodal} and use an element-wise multiplication to combine the latent vector from our advice encoder and the visual feature vector from our visual encoder. 

Formally, our advice LSTM yields a $d$-dimensional latent vector ${\bf{u}}\in\mathcal{R}^{d}$. By combining this vector with the visual feature ${\bf{x}}_{t,i}$ using element-wise multiplication, we obtain a feature vector ${\bf{z}}_{t,i}={\bf{x}}_{t,i}\odot{\bf{u}}$, which is then fed into vehicle controller. Note that vehicle controller takes a new image at every time $t$ (thus, update ${\bf{x}}_{t,i}$) but the latent vector ${\bf{u}}$ remains the same during a period of the event.

Note that we focus on two forms of advice: (i) stimulus-driven and (ii) goal-oriented advice. The former advice (e.g., “watch out a pedestrian”) about perception can be grounded into a context ${\bf{y}}_{t,i}$ via attention maps. We, however, argue that attention maps may not be sufficient to ground the latter advice (e.g., “go straight”), which needs a more direct influence to the controller via an additional element-wise multiplication.

\myparagraph{Synthetic Token.}
We use a synthetic token \texttt{<none>} to indicate unavailable advice input. Since users will not be aware of the full state of the vehicle (they are not driving), the controller should mainly be in charge. Thus, we augment replicate of the dataset that however has a \texttt{<none>} token as the advice input, which exposes the model to events that do not have advice as an input.

\subsection{Interpretable Vehicle Controller}\label{sec:controller}
Providing a controller's internal state is important for advisable systems since it will be used as a ground or an acknowledgment of their advice taken. To this end, we utilize the attention-based driving model~\cite{kim2017interpretable} that provides the controller's internal state by visualizing attention maps -- i.e., where the model visually fixates on image regions that are relevant to the decision.

\myparagraph{Visual Attention.}
Visual attention provides introspective explanations by filtering out non-salient image regions, while image areas inside the attended region have potential causal effect on the output. The goal of visual attention mechanism is to find a context ${\bf{Y}}_t=\{{\bf{y}}_{t,1},{\bf{y}}_{t,2},\dots,{\bf{y}}_{t,l}\}$ by minimizing a loss function, where ${\bf{y}}_{t,i} = \pi(\alpha_{t,i}, {\bf{x}}_{t,i}) = \alpha_{t,i}{\bf{x}}_{t,i}$ for $i=\{1,2,\dots,l\}$. Note that a scalar attention weight value $\alpha_{t,i}$ in $[0,1]$ is associated with a certain grid of input image in such that $\sum_{i}{\alpha_{t,i}}=1$. We use a multi-layer perceptron $f_{\textnormal{attn}}$ to generate $\alpha_{t,i}$, i.e., $\alpha_{t,i}=f_{\textnormal{attn}}({\bf{x}}_{t,i},{\bf{h}}_{t-1})$ conditioned on the previous hidden state ${\bf{h}}_{t-1}$, and the current feature vector ${\bf{x}}_{t,i}$. Softmax regression function is then used to obtain the final attention weight.

\myparagraph{Outputs.}
The outputs of our model are two continuous values of a speed $\hat{v}(t)$ and a steering wheel angle $\hat{s}(t)$. We utilize additional hidden layers $f_{\textnormal{v}}$ and $f_{\textnormal{s}}$, each of which are conditioned on the current hidden state ${\bf{h}}_t$ (of the control LSTM) and a context vector ${\bf{c}}_t$. We generate the context vector by utilizing a function $f_{\textnormal{concat}}$, which concatenates $\{{\bf{c}}_{t,i}\}_{i=1}^{l}=\{{\bf{y}}_{t,i}\odot{\bf{u}}\}_{i=1}^{l}$ to output 1-D vector ${\bf{c}}_t$.

\myparagraph{Internalizing Advice.}\label{sec:internalize}
Stimulus-driven advice provides rich messages about visual saliencies (e.g., traffic lights, pedestrians, signs, etc) that the vehicle controller should typically see these objects while driving. Thus, to internalize such advice, we argue that the driving model must attend to those areas even when such advice is not available. We add a loss term, i.e., the Kullback-Leibler divergence ($D_{KL}$), between two attention maps (i.e., generated with and without advice) to make the driving model refer to the same salient objects:
\begin{equation}
\mathcal{L}_{a} = \lambda_{a}\sum_{t}D_{KL}(\alpha^{w}_{t}||\alpha^{wo}_{t})=\lambda_{a}\sum_{t}\sum_{i=1}^{l}\alpha^{w}_{t,i}(\log\frac{\alpha^{w}_{t,i}}{\alpha^{wo}_{t,i}})
\label{eq:loss_attention}
\end{equation}
where $\alpha^{w}$ and $\alpha^{wo}$ are the attention maps generated by the vehicle controller with and without advice given, respectively. We use a hyperparameter $\lambda_{a}$ to control the strength of the regularization term.

\myparagraph{Loss function.}\label{sec:loss}
Existing models have been trained mainly by minimizing the proportional control error term (i.e., the difference between human-demonstrated and predicted). However, these systems are prone to suffer from two major issues. (i) Oscillation of control predictions -- its prediction has repeated variation against a target value. (ii) Variations in task performance between drivers. 

Inspired by proportional-integral-derivative (PID) controller~\cite{rajamani2011vehicle}, we use the following loss function, which consists of three terms: (i) $\mathcal{L}_{p}$, which is proportional to the error (i.e., $|e_{v}(t)|+|e_{s}(t)|$), where we use the error terms $e_{v}(t)$~=~$v(t)-\hat{v}(t)$ and $e_{s}(t)$~=~$s(t)-\hat{s}(t)$. (ii) $\mathcal{L}_{d}$, which is proportional to the derivative of the error (i.e., $\frac{d}{dt}e_{v}(t)$ and $\frac{d}{dt}e_{s}(t)$), and (iii) $\mathcal{L}_{i}$, which is proportional to the integral of the error, which we use the difference in the vehicle's future course $\theta(t)$ -- a cardinal direction in which a vehicle is to be steered. With the bicycle model assumption~\cite{rajamani2011vehicle} - which assumes that left and right front wheels are represented by one front wheel, we can approximate a steering wheel angle $s_t \approx L/r$, where $L$ is the length of wheelbase and $r$ is the radius of the vehicle's path. Then, we can approximate the vehicle's course $\theta(t)\approx \frac{v(t)\tau}{r}\approx s(t)v(t)$ after the unit time $\tau=1$. Thus, we use the following loss function $\mathcal{L}$:\vspace{-0.6em}
\begin{equation}
\begin{split}
\mathcal{L} = \mathcal{L}_a+\frac{1}{T}\sum_{t=0}^{T-1}\big[\overbrace{|e_{v}(t)|+|e_{s}(t)|}^{\mathcal{L}_p} +\lambda_{i}\overbrace{|\theta(t)-\hat{\theta}(t)|}^{\mathcal{L}_i}\\ +\lambda_{d}\underbrace{\big(|\frac{d}{dt}e_{v}(t)|^2+|\frac{d}{dt}e_{s}(t)|^2\big)}_{\mathcal{L}_d}\big]
\end{split}
\label{eq:loss}
\end{equation}
where $T$ is the number of timesteps. We use hyperparameters $\lambda_{d}$ and $\lambda_{i}$ to control the strength of the terms.

\section{Honda Research Institute-Advice Dataset}
\begin{figure*}[!t]
    \begin{center}
        \includegraphics[width=\linewidth]{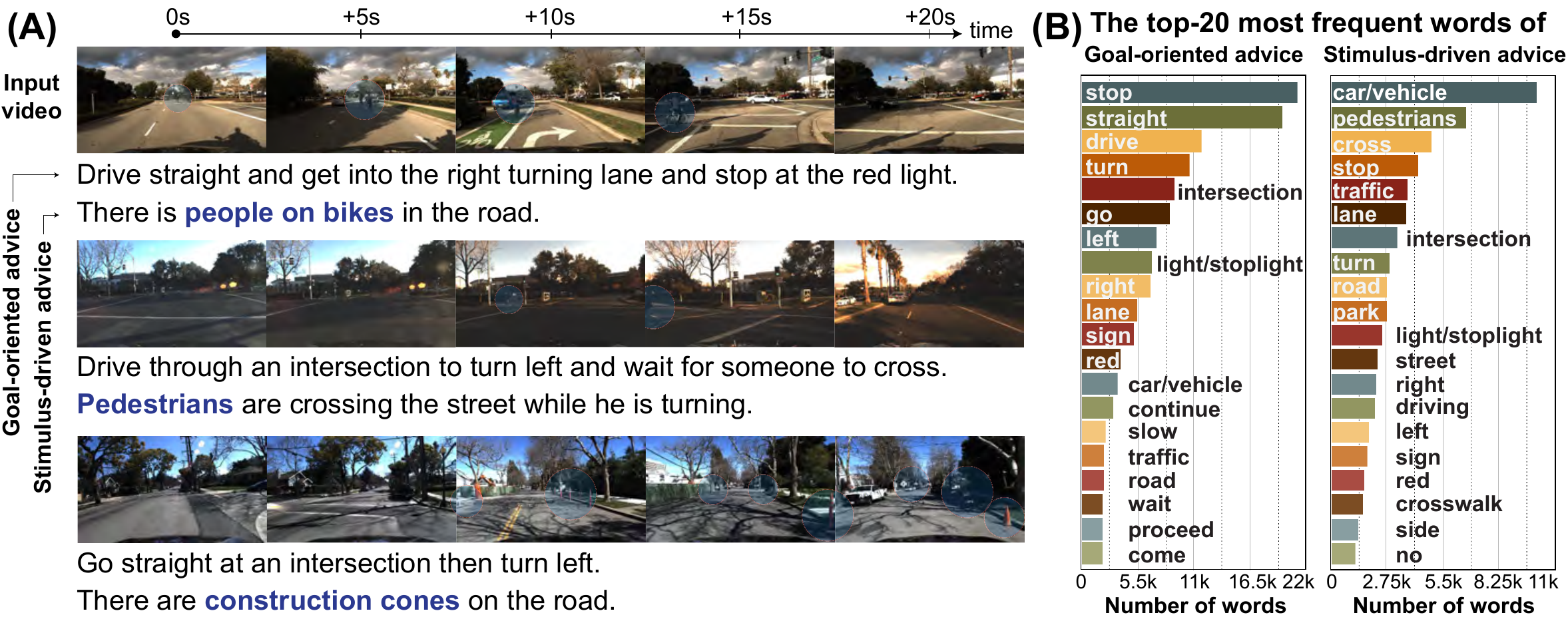}
    \end{center}
    \caption{({\bf{A}}) Examples of input images, which are sampled at every 5 seconds. We also provide examples of the goal-oriented advice and the stimulus-driven advice, which are collected from human annotators followed by a post-processing. We highlight a visual cue (e.g., pedestrians), which are mentioned in advice, with a blue circle on the images. ({\bf{B}}) The counts of top-20 most frequent words used in both types of advice.}
    \label{fig:dataset}\vspace{-0.5em}
\end{figure*}

{
\setlength{\tabcolsep}{4pt}
\renewcommand{\arraystretch}{1.3}
\begin{table}[t]
    \begin{center}
    \caption{Examples of processing annotated descriptions.}
        \label{Table:dataset}
		\resizebox{\linewidth}{!}{%
        \begin{tabular}{@{}lll@{}}
        	\toprule
            Type & Step & Textual Annotation\\
            \midrule
            {\em{action}} desc.&Initial annotation & The driver {\em{went}} straight and {\em{stopped}} at an intersection.\\
            &$\rightarrow$ Present tense& The driver {\em{goes}} straight and {\em{stops}} at an intersection.\\
            &$\rightarrow$ Imperative& {\em{Go}} straight and {\em{stop}} at an intersection.\\
            \midrule
            {\em{attention}} desc.&Initial Annotation & There {\em{was}} a pedestrian pushing a stroller through the crosswalk.\\
            &$\rightarrow$ Present tense& There {\em{is}} a pedestrian pushing a stroller through the crosswalk.\\
            \bottomrule
		\end{tabular}
        }
    \end{center}\vspace{-2em}
\end{table}
}

In order to evaluate the advisable driving model, we have collected Honda Research Institute-Advice Dataset (HAD). In this section, we describe our dataset in terms of the driving videos used to collect human-annotated textual advice, our annotation process, and analysis of the advice collected.

\myparagraph{Driving Videos and Vehicle Control Commands.}
We use 5,675 video clips (over 32 hours), each of which is on average 20 seconds in length. Each video contains around 1-2 driving activities, e.g., passing through an intersection, lane change, stopping, \etc. These videos are randomly collected from a large-scale driving video dataset called HDD~\cite{ramanishka2018toward}. This dataset contains camera videos -- which are captured by a single front-view camera mounted in a fixed position on the roof top of the vehicle. These videos are mostly captured during urban driving near the San Francisco Bay Area, which contain the typical driver's activities (i.e., turning, merging, lane following, etc) on various road types (i.e., highway, residential roads with and without lane markings, etc). Alongside the video data, the dataset provides a set of time-stamped controller area network (CAN) bus records, which contain human driver control inputs (i.e., steering wheel angle).
 
\myparagraph{Annotations.} 
We provide a 20 seconds driving video and ask a human annotator to describe, from a point of view of a driving instructor, what the driver is doing (\emph{action} description for goal-oriented advice) and what the driver should pay attention (\emph{attention} description for stimulus-driven advice). We require that the annotators enter the action description and attention description separately, for example, ``The driver crossed lanes from right to left lane'' and ``There was construction happening on the road'', respectively. Each video clip has 4-5 action descriptions (25,549 in total) and 3-4 attention descriptions (20,080 in total). We then change the descriptions into the present tense (e.g., ``The driver crosses lanes from right to left lane''). Especially for action descriptions, we change them to imperative sentences (e.g., ``Cross lanes from right to left lane''), which are used to offer advice. To ensure the quality of the collected descriptions, we ask another human annotator to proofread the descriptions/advice to correct typographical errors and mistakes in grammar and spelling. In our analysis of annotations, we found that this two-stage annotation is helpful for the annotator to understand the task and perform better. In Figure~\ref{fig:dataset} (A), we provide examples of two types of advice collected along with dashboard camera images (sampled at every 5 seconds).

\myparagraph{Dataset Characteristics.}
Figure~\ref{fig:dataset} (B) shows word counts of the top-20 most frequent words used in the goal-oriented advice and the stimulus-driven advice, respectively. Note that we exclude prepositions, conjunctions, and definite and indefinite articles. Most common goal-oriented advice is related to changes in speed (i.e., stop, slow), driving (i.e., drive, straight, go, etc), and turning (i.e., left, right, turns). Many also include a list of concepts relevant to a driving, such as traffic light/sign, lane, intersection. The stimulus-driven advice covers a diverse list of concepts relevant to the driving scenario, such as the state of traffic/lane, traffic light/sign, pedestrians crossing the street, passing other parked/crossing cars, etc. Although less frequent, some contain references to different types of vehicle (i.e., bus, truck, bike, van, etc), road bumps, and weather conditions.

\section{Experiments}
{
\setlength{\tabcolsep}{4pt}
\renewcommand{\arraystretch}{1.3} 
\begin{table*}[t]
	\begin{center}
	\caption{In order to see the effectiveness of our advice-taking model, we compare the vehicle control prediction performance with other two existing models, which do not take advice (the first two rows). For a fair comparison, we use the identical 5-layer base CNN~\cite{bojarski2016end}. We also share the same input and output layers trained with the same loss function (i.e., the proportional error $\mathcal{L}_p$ alone) used in \cite{kim2017interpretable}. We compare the control prediction performance in terms of three different sets of advice (i.e., the goal-oriented advice (\texttt{ADV}$_G$) only, the stimulus-driven advice ({\texttt{ADV}}$_S$) only, and both). For evaluation, we use the mean of correlation distances (Corr) and the median of absolute errors as well as the 1st (Q1) and 3rd (Q3) quartiles.} 
     \label{Table:vehicleControl}
    	\resizebox{\linewidth}{!}{%
    	\begin{tabular}{@{}llcccccc@{}} \toprule
        	\multirow{2}{*}{Type} &\multirow{2}{*}{Model} & \multicolumn{2}{c}{Advice input} & \multicolumn{2}{c}{Speed (km/h)} & \multicolumn{2}{c}{Steering Wheel Angle (deg)}\\ \cmidrule{3-8}
            & & Training & Testing & Median [Q1, Q3] & Corr & Median [Q1, Q3] & Corr\\ \midrule
            {\em{Non-advisable}}& ConvNet+FF (feed forward network)~\cite{bojarski2016end}& -& -&  6.88 [3.13, 13.1]& .597 & 4.63 [1.80, 12.4] & .366\\
            & ConvNet+LSTM+Attention~\cite{kim2017interpretable} (baseline) & - & - & 3.98 [1.76, 8.10] & .763 & 3.92 [1.54, 10.1] & .469 \\ \midrule
            {\em{Advisable}} & CNN+LSTM+Attention+Advice ({\em{Ours}}) & $\texttt{ADV}_G$ only& $\texttt{ADV}_G$& 4.25 [1.86, 8.46] & .743 & {\bf{3.53 [1.37, 8.83]}} & {\bf{.516}} \\
            & CNN+LSTM+Attention+Advice ({\em{Ours}}) & $\texttt{ADV}_S$ only& $\texttt{ADV}_S$& {\bf{3.28 [1.47, 6.46]}} & {\bf{.782}} & 3.78 [1.45, 9.93] & .484 \\ \cmidrule{2-8}
            & CNN+LSTM+Attention+Advice ({\em{Ours}}) & $\texttt{ADV}_G$+$\texttt{ADV}_S$& $\texttt{ADV}_G$ & 3.78 [1.67, 7.50] & .763 & 3.54 [1.36, 9.21] & .512 \\
            & CNN+LSTM+Attention+Advice ({\em{Ours}}) & $\texttt{ADV}_G$+$\texttt{ADV}_S$& $\texttt{ADV}_S$& 3.78 [1.68, 7.46] & .763 & 3.78 [1.41, 9.51] & .511 \\
            \bottomrule
        \end{tabular}}
     \end{center}\vspace{-1em}
\end{table*}
}

\myparagraph{Training and Evaluation Details.}
We use a single LSTM layer for all the components of our framework. Our model is trained end-to-end using random initialization (i.e., no pre-trained weights). For training, we use Adam optimization algorithm~\cite{kingma2014adam} and dropout~\cite{srivastava2014dropout} of 0.5 at hidden state connections and Xavier initialization~\cite{glorot2010understanding}. Our model takes 1-3 days (depending on types of CNN used) to train and can process over 100 frames on average per second on a single Titan Xp GPU. We use two mathematical criteria (the statistics of absolute errors and the correlation distance) to quantitatively evaluate their performance by comparing with ground-truth human-demonstrated control commands.

\myparagraph{Advisable vs. Non-advisable models.}
As shown in Table~\ref{Table:vehicleControl}, we first compare the vehicle control prediction performance to see our advice-taking driving model can outperform other existing driving models that do not take advice. To this end, we implemented two other existing models, i.e., (1) CNN+FF (Feed forward network)~\cite{bojarski2016end} and (2) CNN+LSTM+Attention~\cite{kim2017interpretable}. For a fair comparison, all models used the identical 5-layer CNN~\cite{bojarski2016end} as the convolutional (visual) feature encoder trained by minimizing the loss term $\mathcal{L}_p$ only (same as used in \cite{kim2017interpretable}. See Equation~\ref{eq:loss}). This visual encoder produces a 12$\times$20$\times$64-dimensional feature cube from the last convolutional layer. In the later section, we will also explore further potential performance improvements with more expressive neural networks over this base CNN configuration. 

\begin{figure*}[!t]
    \begin{center}
        \includegraphics[width=\linewidth]{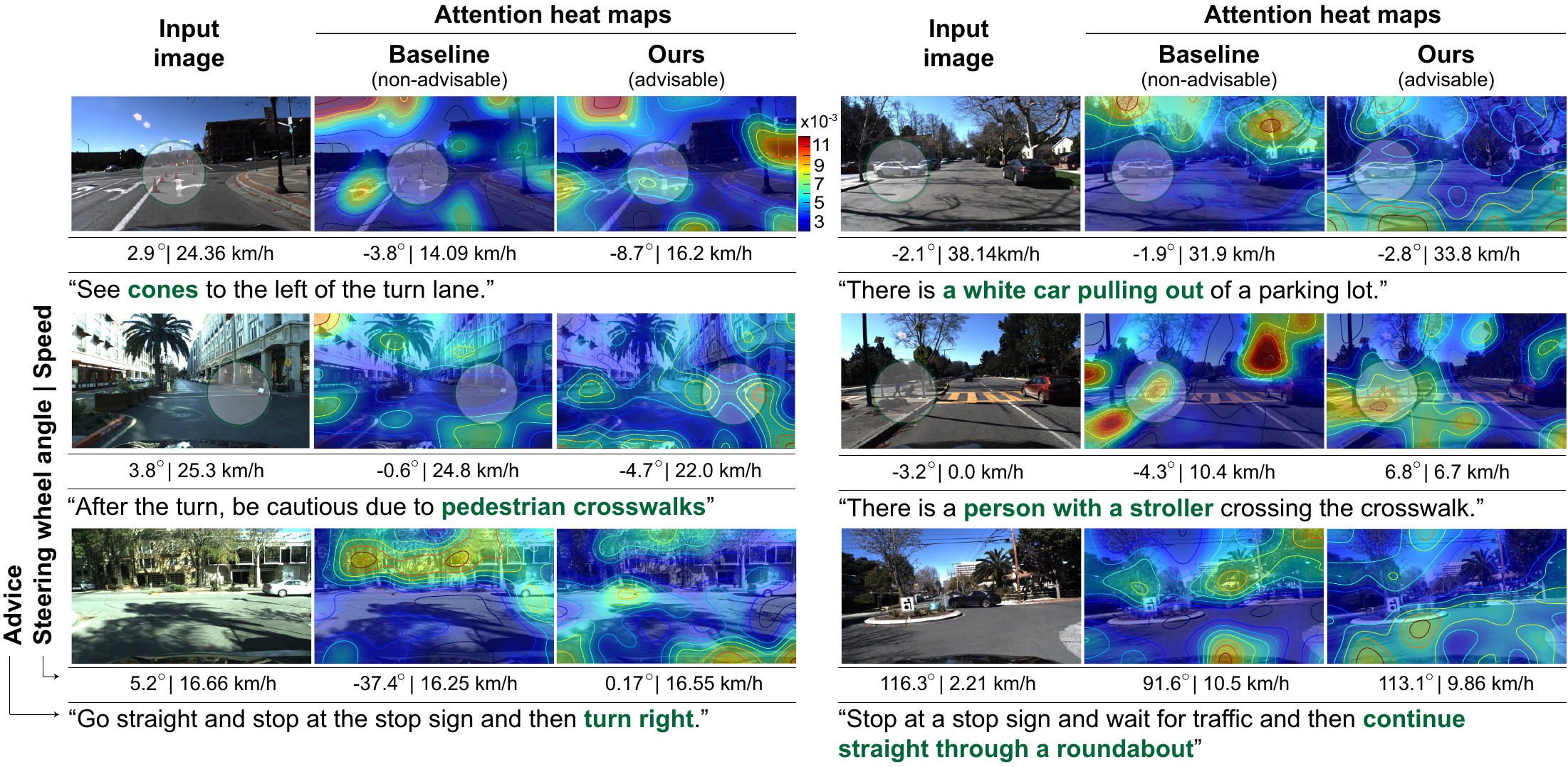}
    \end{center}
    \caption{Attention heat maps comparison. We provide input raw images and attention heat maps generated by the existing attention-based driving model~\cite{kim2017interpretable} (Baseline column), and our model trained with all types of advice together (Ours column). We highlight key object-centric words (as appropriate row 1 \& 2), e.g., cones and a white car pulling out, in green as well as corresponding salient objects in a green circle overlaid on images.}
    \label{fig:qualitative}
\end{figure*}

\begin{figure*}[!t]
    \begin{center}
        \includegraphics[width=\linewidth]{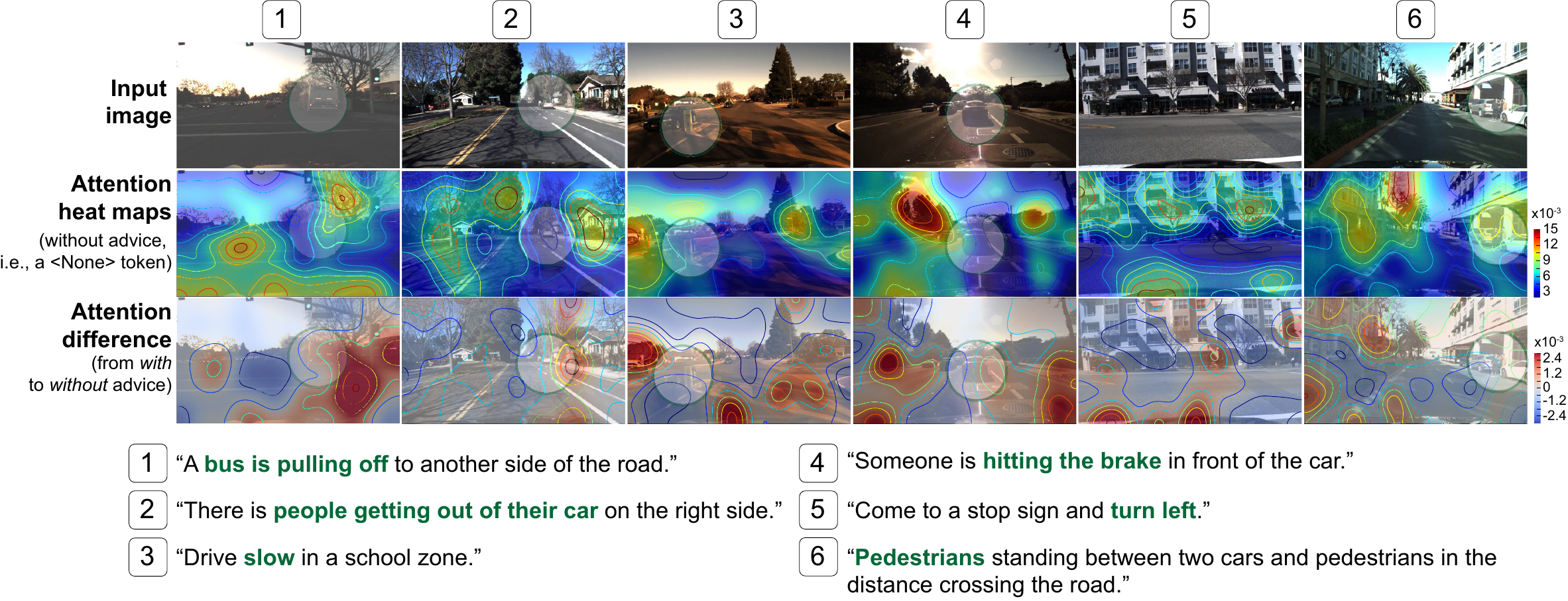}
    \end{center}
    \caption{We compared attention heat maps generated {\em{with}} and {\em{without}} advice as an input in testing time. We visualize raw input images with salient objects marked by a green circle, e.g., a bus pulling off, which is mentioned by an advice input (1st row). The provided advice (1-6) is provided at the bottom of the figure. We visualize attention heat maps from our trained model but with a synthetic token \texttt{<none>} (i.e., without advice, 2nd row). Attention map differences between those with and without advice (3rd row), where red parts indicate where the model (with advice) pays more attention.}
    \label{fig:acknowledge}\vspace{-1em}
\end{figure*}

In Table~\ref{Table:vehicleControl}, we report a summary of our experiments validating the quantitative effectiveness of our advice-taking approach. Comparing with the non-advisable models (rows 1-2), our advisable models all gave better scores for vehicle control prediction. As we will see in the next section, we observe that our advisable driving model focuses more on driving-related objects (whether provided as advice or not) than others that do not take advice during training and testing phases. For example, in Figure~\ref{fig:qualitative} and~\ref{fig:acknowledge}, our advisable model pays more attention to pedestrians crossing, a car pulling out, and construction cones. More importantly, advice like ``{\em{stop at a stop sign}}'' or ``{\em{there is a person with a stroller crossing the crosswalk}}'' may reflect typical links between visual causes and actions of human driver behavior. The data suggests that taking advice in controller helps imitate more closely human driver behaviors. Biasing the controller by taking advice improves the plausibility of its output from a human perspective.

\myparagraph{Types of Advice Matter.}
We further examine the performance comparison with two different types of advice: the goal-oriented advice (e.g., ``stop at the intersection'') and the stimulus-driven advice (e.g., ``there is a pedestrian crossing''). In Table~\ref{Table:vehicleControl} (rows 3-4), we report vehicle control prediction accuracy when each of which types of advice is given to the model. In our analysis, the goal-oriented advice provides better control accuracy for predicting steering wheel angle commands. This is mainly due to the fact that the goal-oriented advice conveys the more direct messages, which may include navigational command on how the vehicle behaves (e.g., go/stop and turn). The stimulus-driven advice, which conveys rich messages about visual saliencies (e.g., red light, stop sign, and intersection), provides better predicting accuracy for vehicle's speed prediction.

{
\setlength{\tabcolsep}{4pt}
\renewcommand{\arraystretch}{1.3} 
\begin{table}[t]
	\begin{center}
	\caption{Recall from Section~\ref{sec:internalize}, we propose an advice internalization technique -- which minimizes the difference between two attention maps (generated {\em{with}} and {\em{without}} advice inputs) and thus makes the driving model refer to the same salient objects. Note that we use a synthetic token \texttt{<none>} to indicate when advice inputs are not available. We used $\lambda_a$ as 50 (by the grid-search method).
	} 
    \label{Table:internalize}
    	\resizebox{\linewidth}{!}{%
    	\begin{tabular}{@{}lcccccc@{}} \toprule
        	\multirow{2}{*}{Model} & \multicolumn{2}{c}{Advice input} & \multicolumn{2}{c}{Speed (km/h)} & \multicolumn{2}{c}{Steering Wheel Angle (deg)}\\ \cmidrule{2-7}
            & Training & Testing & Median [Q1, Q3] & Corr & Median [Q1, Q3] & Corr\\ \midrule
            {\em{no}} advice internalization & \texttt{ADV}$_S$ & \texttt{<None>} & 3.55 [1.58, 7.12] & .777 & 4.01 [1.59, 10.1] & .479 \\ 
            {\em{w/}} advice internalization & \texttt{ADV}$_S$ & \texttt{<None>} & {\bf{3.36 [1.51, 6.62]}} & {\bf{.784}} & {\bf{3.96 [1.55, 10.0]}} & {\bf{.480}} \\ 
            \bottomrule
        \end{tabular}}
    \end{center}\vspace{-2em}
\end{table}
}

\myparagraph{Qualitative Analysis of Attention Maps.}
As shown in Figure~\ref{fig:qualitative}, we qualitatively compared with our baseline by visualizing attention heat maps - the highlighted image region has a potential influence on the network's outputs. While all models see driving-related common visual cues (i.e., lane markings), we observed that our advice-taking model focuses more on both advice-related cues (i.e., pedestrian crossing, construction cones, a car pulling out, etc) or visual objects relevant to the certain driving scenario (i.e., vehicles, crosswalk, pedestrians, etc).

\myparagraph{Internalizing Advice Taken.}
Users will not usually be aware of the full state of the vehicle (they are not driving), the vehicle controller should mostly be in charge and the human-to-vehicle advice might occasionally be unavailable. As summarized in Table~\ref{Table:internalize}, we further examine the performance comparison with no advice available (we use a synthetic token \texttt{<none>} to indicate unavailable advice input) in a testing time. Interestingly, we observe that (i) the performance of a model trained with the stimulus-driven advice is not degraded much whenever advice inputs are not available in testing (its control performance is still better than other non-advisable approaches), (ii) our advice internalization technique (see Equation~\ref{eq:loss_attention}) further improves the control performance toward those having advice inputs.

In Figure~\ref{fig:acknowledge}, we further examine the effect of advice internalization by visualizing attention heat maps. We first visualize attention maps generated with no advice provided (i.e., using a \texttt{<none>} token, see middle row). Then, we visualize the attention map changes when the model takes ground-truth advice as an input (see bottom row). Our result reveals that our model is still able to see driving-related visual cues (i.e., traffic lights or lanes), whereas advice inputs can bias the model to refer to objects, which is related to the advice given.

{
\setlength{\tabcolsep}{4pt}
\renewcommand{\arraystretch}{1.3} 
\begin{table}[t]
	\begin{center}
	\caption{We compared the vehicle control prediction performance with four different visual encoders. Except for the visual encoder part, we use the same training strategy.} 
    \label{Table:visualencoder}
    	\resizebox{\linewidth}{!}{%
    	\begin{tabular}{@{}lcccc@{}} \toprule
        	\multirow{2}{*}{CNN base} & \multicolumn{2}{c}{Speed (km/h)} & \multicolumn{2}{c}{Steering Wheel Angle (deg)}\\ \cmidrule{2-5}
            &Median [Q1, Q3] & Corr & Median [Q1, Q3] & Corr\\ \midrule
            MobileNet~\cite{howard2017mobilenets} & 3.93 [1.73, 7.80] & .753 & 4.20 [1.65, 10.7] & .463 \\
            Bojarski~\etal~\cite{bojarski2016end} & 3.78 [1.68, 7.49] & .763 & 3.58 [1.39, 9.34] & .512 \\
            Inception v3~\cite{szegedy2016rethinking} & {\bf{2.89 [1.31, 5.59]}} & .795 & {\bf{3.47 [1.34, 8.76]}} & {\bf{.525}} \\
            Inception-ResNet-v2~\cite{szegedy2017inception} & 2.93 [1.33, 5.63] & {\bf{.796}} & 3.54 [1.36, 9.19] & .491 \\
            \bottomrule
        \end{tabular}}
    \end{center}\vspace{-2em}
\end{table}
}

\myparagraph{Visual Encoder Comparison.}
We further examine variants of our proposed model using four different widely-used visual feature encoders. We used the output of intermediate layers from Bojarski~\etal~\cite{bojarski2016end}, Inception v3~\cite{szegedy2016rethinking}, MobileNet~\cite{howard2017mobilenets}, and Inception-ResNet-v2~\cite{szegedy2017inception}. We trained all models in an end-to-end manner using random initialization, and we used both types of advice as an input in the training and testing phases (averaged scores are reported). As reported in Table~\ref{Table:visualencoder}, the result reveals that control prediction accuracy can be generally expected to improve when using a deeper CNN architecture, which learns more expressive visual features. Visual features from the Inception v3-based architecture lead the best performance improvement against other three architectures.

\begin{figure}[!t]
    \begin{center}
        \includegraphics[width=\linewidth]{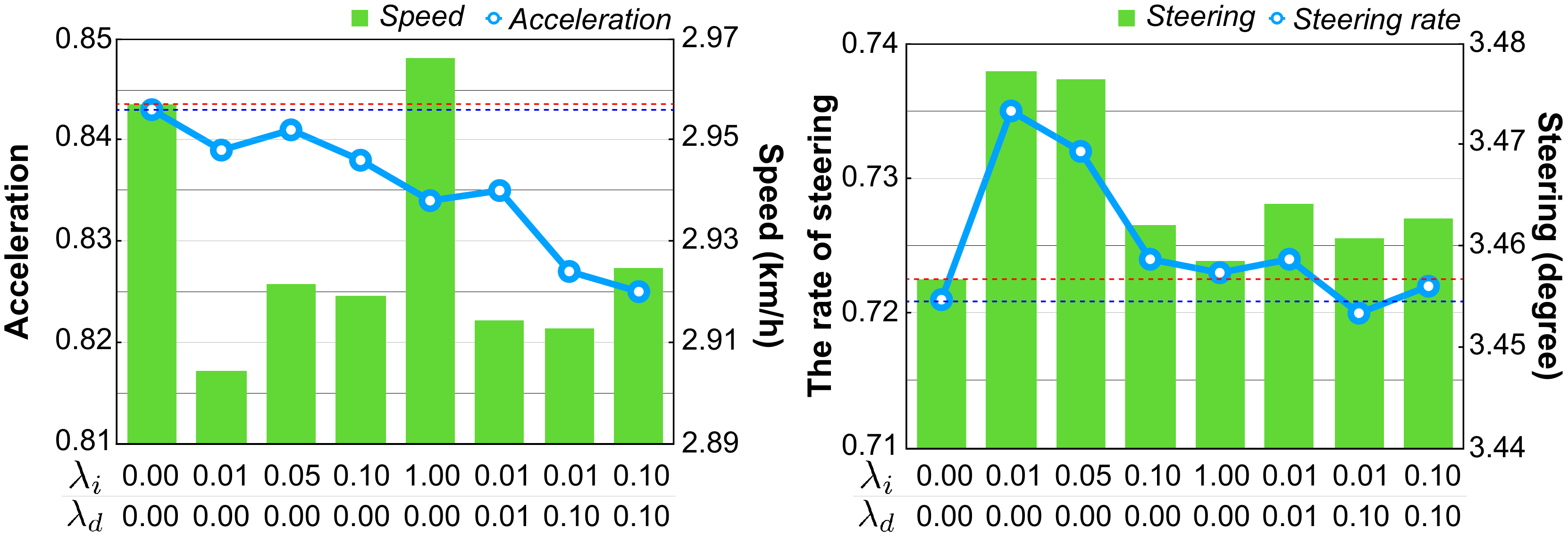}
    \end{center}
    \caption{Control performance comparison with different sets of hyperparameters ($\lambda_{d}$, $\lambda_{i}$) (see Equation~\ref{eq:loss}). Along with proportional prediction errors (green bar), we also visualize derivative errors (blue line). We report the median value of absolute error.}\vspace{-1em}
    \label{fig:lossft}
\end{figure}

\myparagraph{Effect of Regularization.}
Recall from Section~\ref{sec:loss}, we explored the loss function $\mathcal{L}$, which contains three terms -- $\mathcal{L}_{p}$ (proportional error), $\mathcal{L}_{d}$ (derivative error), and $\mathcal{L}_{i}$ (integral error). We use two hyperparameters $\lambda_{d}$ and $\lambda_{i}$ to control the strength of the corresponding terms. Figure~\ref{fig:lossft} shows control command prediction errors with different combinations of hyperparameters in terms of the median value of absolute errors. We also visualize the error of the acceleration (the derivative of speed) and the steering angle rate (the derivative of steering angle command). The impact of adding these loss terms is dominant in the prediction of speed, whereas the performance in steering is slightly degraded. We obtained marginal improvement by adding integral loss term ($\lambda_{i}$) in speed predictions, while derivative errors are reduced by adding derivative loss term ($\lambda_{d}$).

\section{Conclusion}
We described an advisable driving model for self-driving cars by incorporating a textual encoder that understands human-to-vehicle natural language advice and grounds it into the vehicle controller. We showed that (i) taking advice improves vehicle control prediction accuracy compared to baselines, (ii) our advice-taking model really sees advice-related visual cues and such advice can be internalized, and (iii) our Honda Research Institute-Advice Dataset (HAD) allows us to train and evaluate our advisable model and we will make the dataset available upon publication.

This is a first paper on the use of advice, but this design is most appropriate for turn-by-turn (short duration) advice. Since our data comprised short clips, advice was effective throughout the clip. It will be worth exploring other styles of advice, such as per-ride advice (gentle, fast, etc) and rule-based global advice. 

\vspace{-0.5em}
\paragraph{Acknowledgements.} This work was part of J. Kim's summer/fall internship at Honda Research Institute, USA and also supported by DARPA XAI program and Berkeley DeepDrive.

{\small
\bibliographystyle{ieee}
\bibliography{egbib}

\begin{thebibliography}{10}\itemsep=-1pt

\bibitem{artzi2013weakly}
Y.~Artzi and L.~Zettlemoyer.
\newblock Weakly supervised learning of semantic parsers for mapping
  instructions to actions.
\newblock {\em Transactions of the Association of Computational Linguistics},
  1:49--62, 2013.

\bibitem{bojarski2016visualbackprop}
M.~Bojarski, A.~Choromanska, K.~Choromanski, B.~Firner, L.~Jackel, U.~Muller,
  and K.~Zieba.
\newblock Visualbackprop: visualizing cnns for autonomous driving.
\newblock {\em arXiv preprint}, 2016.

\bibitem{bojarski2016end}
M.~Bojarski, D.~Del~Testa, D.~Dworakowski, B.~Firner, B.~Flepp, P.~Goyal, L.~D.
  Jackel, M.~Monfort, U.~Muller, J.~Zhang, et~al.
\newblock End to end learning for self-driving cars.
\newblock {\em CoRR abs/1604.07316}, 2016.

\bibitem{chi2017learning}
L.~Chi and Y.~Mu.
\newblock Learning end-to-end autonomous steering model from spatial and
  temporal visual cues.
\newblock In {\em Proceedings of the Workshop on Visual Analysis in Smart and
  Connected Communities}, pages 9--16. ACM, 2017.

\bibitem{fernando2017going}
T.~Fernando, S.~Denman, S.~Sridharan, and C.~Fookes.
\newblock Going deeper: Autonomous steering with neural memory networks.
\newblock In {\em Computer Vision Workshop (ICCVW), 2017 IEEE International
  Conference on}, pages 214--221. IEEE, 2017.

\bibitem{glorot2010understanding}
X.~Glorot and Y.~Bengio.
\newblock Understanding the difficulty of training deep feedforward neural
  networks.
\newblock In {\em Proceedings of International Conference on Artificial
  Intelligence and Statistics (AISTATS)}, volume~9, pages 249--256, 2010.

\bibitem{howard2017mobilenets}
A.~G. Howard, M.~Zhu, B.~Chen, D.~Kalenichenko, W.~Wang, T.~Weyand,
  M.~Andreetto, and H.~Adam.
\newblock Mobilenets: Efficient convolutional neural networks for mobile vision
  applications.
\newblock {\em arXiv preprint arXiv:1704.04861}, 2017.

\bibitem{kim2017interpretable}
J.~Kim and J.~Canny.
\newblock Interpretable learning for self-driving cars by visualizing causal
  attention.
\newblock {\em Proceedings of the IEEE International Conference on Computer
  Vision (ICCV)}, pages 2942--2950, 2017.

\bibitem{kim2018textual}
J.~Kim, A.~Rohrbach, T.~Darrell, J.~Canny, and Z.~Akata.
\newblock Textual explanations for self-driving vehicles.
\newblock In {\em Proceedings of European Conference on Computer Vision
  (ECCV)}, pages 577--593. Springer, Cham, 2018.

\bibitem{kingma2014adam}
D.~Kingma and J.~Ba.
\newblock Adam: A method for stochastic optimization.
\newblock {\em Proceedings of International Conference for Learning
  Representations (ICLR)}, 2015.

\bibitem{kuhlmann2004guiding}
G.~Kuhlmann, P.~Stone, R.~Mooney, and J.~Shavlik.
\newblock Guiding a reinforcement learner with natural language advice: Initial
  results in robocup soccer.
\newblock In {\em The AAAI-2004 workshop on supervisory control of learning and
  adaptive systems}. San Jose, CA, 2004.

\bibitem{li2016dialogue}
J.~Li, A.~H. Miller, S.~Chopra, M.~Ranzato, and J.~Weston.
\newblock Dialogue learning with human-in-the-loop.
\newblock {\em arXiv preprint arXiv:1611.09823}, 2016.

\bibitem{ling2017teaching}
H.~Ling and S.~Fidler.
\newblock Teaching machines to describe images via natural language feedback.
\newblock In {\em Proceedings of the International Conference on Neural
  Information Processing Systems}, pages 5075--5085, 2017.

\bibitem{liu2016ssd}
W.~Liu, D.~Anguelov, D.~Erhan, C.~Szegedy, S.~Reed, C.-Y. Fu, and A.~C. Berg.
\newblock Ssd: Single shot multibox detector.
\newblock In {\em Proceedings of European Conference on Computer Vision
  (ECCV)}, pages 21--37. Springer, 2016.

\bibitem{mccarthy1960programs}
J.~McCarthy.
\newblock {\em Programs with common sense}.
\newblock RLE and MIT computation center, 1960.

\bibitem{misra2016tell}
D.~K. Misra, J.~Sung, K.~Lee, and A.~Saxena.
\newblock Tell me dave: Context-sensitive grounding of natural language to
  manipulation instructions.
\newblock {\em The International Journal of Robotics Research},
  35(1-3):281--300, 2016.

\bibitem{misra2015environment}
D.~K. Misra, K.~Tao, P.~Liang, and A.~Saxena.
\newblock Environment-driven lexicon induction for high-level instructions.
\newblock In {\em Proceedings of the 53rd Annual Meeting of the Association for
  Computational Linguistics and the 7th International Joint Conference on
  Natural Language Processing (Volume 1: Long Papers)}, volume~1, pages
  992--1002, 2015.

\bibitem{park2018multimodal}
D.~H. Park, L.~A. Hendricks, Z.~Akata, A.~Rohrbach, B.~Schiele, T.~Darrell, and
  M.~Rohrbach.
\newblock Multimodal explanations: Justifying decisions and pointing to the
  evidence.
\newblock In {\em Proceedings of the IEEE Conference on Computer Vision and
  Pattern Recognition (CVPR)}, 2018.

\bibitem{rajamani2011vehicle}
R.~Rajamani.
\newblock {\em Vehicle dynamics and control}.
\newblock Springer Science \& Business Media, 2011.

\bibitem{ramanishka2018toward}
V.~Ramanishka, Y.-T. Chen, T.~Misu, and K.~Saenko.
\newblock Toward driving scene understanding: A dataset for learning driver
  behavior and causal reasoning.
\newblock In {\em Proceedings of the IEEE Conference on Computer Vision and
  Pattern Recognition (CVPR)}, pages 7699--7707, 2018.

\bibitem{srivastava2014dropout}
N.~Srivastava, G.~E. Hinton, A.~Krizhevsky, I.~Sutskever, and R.~Salakhutdinov.
\newblock Dropout: a simple way to prevent neural networks from overfitting.
\newblock {\em Journal of Machine Learning Research}, 15(1):1929--1958, 2014.

\bibitem{szegedy2017inception}
C.~Szegedy, S.~Ioffe, V.~Vanhoucke, and A.~A. Alemi.
\newblock Inception-v4, inception-resnet and the impact of residual connections
  on learning.
\newblock In {\em AAAI}, volume~4, page~12, 2017.

\bibitem{szegedy2016rethinking}
C.~Szegedy, V.~Vanhoucke, S.~Ioffe, J.~Shlens, and Z.~Wojna.
\newblock Rethinking the inception architecture for computer vision.
\newblock In {\em Proceedings of the IEEE Conference on Computer Vision and
  Pattern Recognition (CVPR)}, pages 2818--2826, 2016.

\bibitem{tellex2011understanding}
S.~Tellex, T.~Kollar, S.~Dickerson, M.~R. Walter, A.~G. Banerjee, S.~J. Teller,
  and N.~Roy.
\newblock Understanding natural language commands for robotic navigation and
  mobile manipulation.
\newblock In {\em AAAI}, volume~1, page~2, 2011.

\bibitem{tung2018reward}
H.-Y.~F. Tung, A.~W. Harley, L.-K. Huang, and K.~Fragkiadaki.
\newblock Reward learning from narrated demonstrations.
\newblock {\em Proceedings of the IEEE Conference on Computer Vision and
  Pattern Recognition (CVPR)}, 2018.

\bibitem{weston2016dialog}
J.~E. Weston.
\newblock Dialog-based language learning.
\newblock In {\em Advances in Neural Information Processing Systems}, pages
  829--837, 2016.

\bibitem{xu2016end}
H.~Xu, Y.~Gao, F.~Yu, and T.~Darrell.
\newblock End-to-end learning of driving models from large-scale video
  datasets.
\newblock In {\em Proceedings of the IEEE Conference on Computer Vision and
  Pattern Recognition (CVPR)}, pages 2174--2182, 2017.

\end{thebibliography}
}

\end{document}